\documentclass[letterpaper]{article} 
\usepackage{aaai25}  
\usepackage{times}  
\usepackage{helvet}  
\usepackage{courier}  
\usepackage[hyphens]{url}  
\usepackage{graphicx} 
\usepackage{natbib}  
\usepackage{caption} 
\usepackage{booktabs}
\usepackage{algorithm}
\usepackage{algorithmic}
\usepackage{amssymb}
\usepackage{amsmath}

\usepackage{newfloat}
\usepackage{listings}
\DeclareCaptionStyle{ruled}{labelfont=normalfont,labelsep=colon,strut=off} 
\floatstyle{ruled}
\newfloat{listing}{tb}{lst}{}
\floatname{listing}{Listing}
\title{FLAMe: Federated Learning with Attention Mechanism using Spatio-Temporal Keypoint Transformers for Pedestrian Fall Detection in Smart Cities}
\author {
    Byeonghun Kim\textsuperscript{\rm 1},
    Byeongjoon Noh\textsuperscript{\rm 2}\thanks{Corresponding author}
}
\affiliations {
    \textsuperscript{\rm 1}Department of Future Convergence Technology, Soonchunhyang University, Asan, 31538, South Korea\\
    \textsuperscript{\rm 2}Department of AI and Big Data, Soonchunhyang University, Asan, 31538, South Korea\\
    \{byeonghun, powernoh\}@sch.ac.kr
}

\usepackage{bibentry}

\begin{document}

\maketitle

\begin{abstract}
In smart cities, detecting pedestrian falls is a major challenge to ensure the safety and quality of life of citizens. In this study, we propose a novel fall detection system using FLAMe (Federated Learning with Attention Mechanism), a federated learning (FL) based algorithm. FLAMe trains around important keypoint information and only transmits the trained important weights to the server, reducing communication costs and preserving data privacy. Furthermore, the lightweight keypoint transformer model is integrated into the FL framework to effectively learn spatio-temporal features. We validated the experiment using 22,672 video samples from the ``Fall Accident Risk Behavior Video-Sensor Pair data'' dataset from AI-Hub. As a result of the experiment, the FLAMe-based system achieved an accuracy of 94.02\% with about 190,000 transmission parameters, maintaining performance similar to that of existing centralized learning while maximizing efficiency by reducing communication costs by about 40\% compared to the existing FL algorithm, FedAvg. Therefore, the FLAMe algorithm has demonstrated that it provides robust performance in the distributed environment of smart cities and is a practical and effective solution for public safety.
\end{abstract}

%

\section{Introduction}

\begin{figure*}[t]
    \centering
    \includegraphics[width=0.9\textwidth]{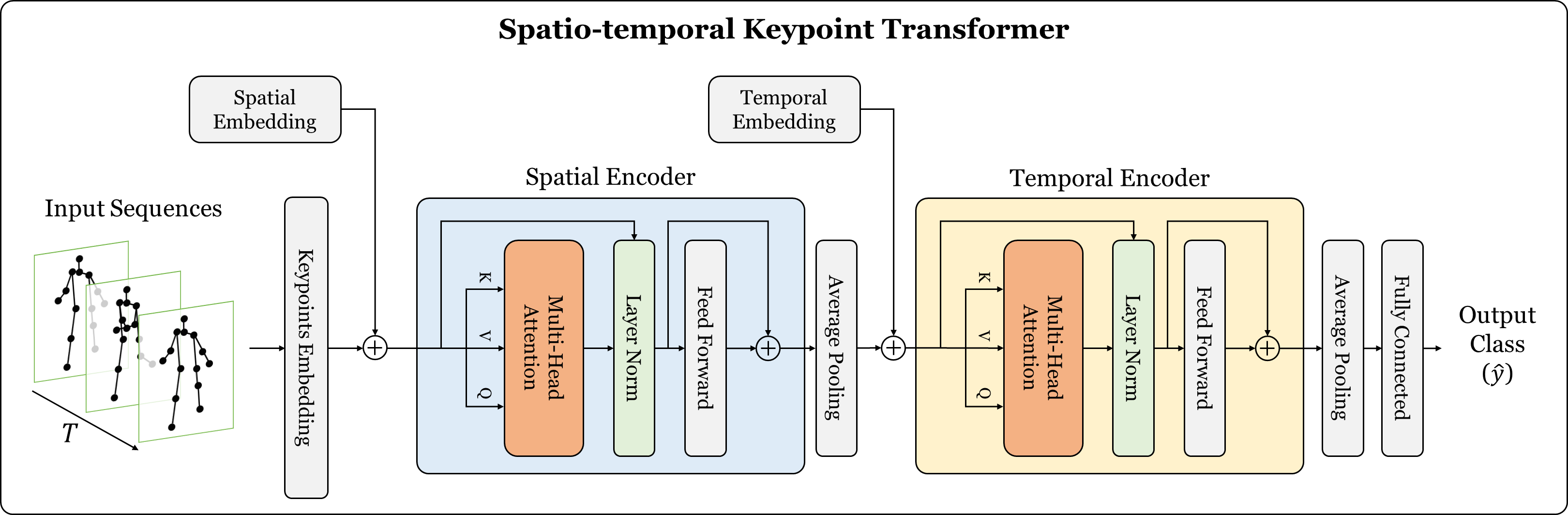}
    \caption{Architecture of the spatio-temporal keypoint transformer model}
    \label{fig:architecture}
\end{figure*}

Smart cities are a paradigm of urban operations that leverage intelligent technologies to improve the safety and quality of life for citizens, and traffic safety systems are one of its key components \cite{hassan2023intelligent, adewopo2023review}. In particular, pedestrian falls are a major safety challenges, which requires the development of technologies to detect and respond to them rapidly \cite{baek2024unveiling, world2021step}. Traditionally, AI and deep learning-based fall detection technologies and closed-circuit television (CCTV) have been utilized to detect pedestrian falls, however, these approaches have limitations such as privacy leakage, network communication bottlenecks, and performance degradation due to environmental conditions such as lighting changes or camera angles \cite{inturi2023novel, amsaprabhaa2023multimodal}. These challenges significantly restrict the practical application of fall detection techniques in the complex environments of smart cities.

Federated learning (FL) provides a promising solution to address these issues in urban CCTV systems \cite{utomo2023federated, mcmahan2017communication}. FL allows data privacy and relieves communication bottlenecks by sharing only model parameters with a central server while preserving data on local devices \cite{wen2023survey, qi2023fl, al2023federated}. However, traditional FL algorithms suffer from the high dimensionality and computational cost of CCTV video data, which makes efficient learning impossible in resource-constrained environments. In addition, they aggregate all parameters equally without considering the model structure, resulting in decreased learning performance and unnecessary communication costs. To solve this problem, an efficient aggregation strategy that reflects the model structure and data characteristics to focus on the important information is required. Recently, the authors in \cite{gahlan2024aflemp} proposed a method to improve the learning efficiency and data heterogeneity challenges by integrating a attention mechanism within the FL framework. However, it still aggregates all parameters equally without considering the model structure, which results in unnecessary communication costs.

Studies have investigated various computer-vision techniques to enhance video-based fall-detection accuracy \cite{wu2023video, wu2023robust}. Conventional methods extract frames from the video and utilize convolutional neural network (CNN) and long short-term memory (LSTM) networks to detect falls \cite{bui2024elderly, inturi2023novel, gomes2022multi}. However, these methods are susceptible to environmental factors such as occlusion and lighting changes, and suffer from computational cost as the number of channels increases. To solve this challenge, models such as graph convolutional network (GCN) and ARFDNet utilizing keypoint data have been proposed \cite{yang2024sma, fei2023flow, yadav2022arfdnet}, demonstrating their strengths in data lightweight and reduced environmental sensitivity. In particular, transformer models based on keypoint data effectively learn spatio-temporal features through an attention mechanism, accurately detect pedestrian movements in complex environments, and provide high computational efficiency and stability. Thus, we implemented a fall-detection model using a keypoint transformer-based network.

In this study, we propose a pedestrian fall detection system based on the \textbf{F}ederated \textbf{L}earning with \textbf{A}ttention \textbf{Me}chanism (\textbf{FLAMe}) algorithm, which effectively aggregates keypoint information by utilizing the attention mechanism, according to the structure and data characteristics of keypoint transformers. FLAMe is designed to integrate the attention mechanism with the FL framework to efficiently learn around important keypoint features. This allows it to avoid unnecessary parameter updates and share parameters only for important information, significantly reducing communication costs and increasing efficiency. The system operates as follows: First, each CCTV operates as a client, processing the video data to extract human keypoints and identify key spatio-temporal features. Second, each local client trains a keypoint transformer model using local datasets, and utilizes a multi-head attention mechanism to effectively extract spatio-temporal features. Finally, among the learned parameters, only the important keypoint parameters are sent to the central server, which aggregates them to generate a global model. This provides a practical and effective pedestrian fall detection solution in the complex environment of smart cities.

The contributions of this study can be summarized as follows: 1) We developed a novel FLAMe algorithm that efficiently learns by focusing on important keypoints, reflecting the complex environment of smart cities and various CCTV conditions. 2) We proposed a lightweight keypoint transformer model that improves computational efficiency and effectively learns spatio-temporal features. Furthermore, the proposed fall detection model significantly reduces the parameters compared to the widely used baseline model, which significantly reduces the consumption of memory and computational resources and achieves lightweighting. 3) We protect pedestrian privacy by using the FL-based approach, and can provide stable and high performance in various environments while reducing communication cost by sharing parameters focused on important keypoints. Furthermore, we validated the feasibility and applicability of the proposed system using a fall detection dataset: Fall Accident Risk Behavior Video-Sensor Pair data \cite{AIHubdata}.

\section{Methodology}
In this section, we present the design and main components of a pedestrian fall detection system with practical application in a smart city environment. The proposed system is based on the FLAMe algorithm, which is designed to maximize communication efficiency even in resource-constrained environments, highlighting the practical applicability of associative learning. In addition, a lightweight spatio-temporal keypoint transformer model designed for practical smart city applications enables effective learning of spatio-temporal features while the data processing and computational costs are minimized. The system mainly consists of 1) a preprocessing process for local dataset generation, 2) a lightweight spatio-temporal keypoint transformer design, and 3) the FLAMe algorithm based on the attention mechanism.

\subsection{Preprocessing for local dataset}
Each client performs preprocessing independently without uploading data to the server in order to build a local dataset. In this study, we use you look only once version 8 pose (YOLOv8-pose) \cite{YOLOv8}, which is widely used in the computer vision field, to estimate the pose of pedestrians. The YOLOv8-pose model integrates object detection and pose estimation, allowing pose estimation in single stage, and provides efficient and reliable results. First, the model processes the input frame to generate a bounding box and keypoint coordinates $(x, y, confidence)$ for each pedestrian. The YOLOv8-pose estimator used is a pre-trained model based on the Microsoft Common Objects in Context (COCO) image dataset \cite{MSCOCO}, which is capable of performing accurate 2D pose estimation without further training when a person is the main object. In addition, to uniform the length of the input data, we sampled 45 of the frames of different lengths from the extracted video to create a standardized dataset.

\subsection{Spatio-temporal keypoint transformer}
Fig.\ref{fig:architecture} illustrates the architecture of the local model, which employs two transformer encoders to extract spatio-temporal features from keypoint coordinates. The input data is a keypoint sequence $X \in \mathbb{R}^{T \times N \times 2}$, where $T$ is the number of sequences, and each sequence contains $N$ keypoints $(x, y)$. The sequence $X$ is linearly transformed into an embedding space with dimension $d$, resulting in $z_e \in \mathbb{R}^{T \times N \times d}$. A spatial positional embedding $p_s \in \mathbb{R}^{N \times d}$ is then added to preserve the spatial location of each keypoint, forming the input representation $h_s = z_e \bigoplus p_s$.

The spatial transformer encoder utilizes a multi-head attention mechanism to model interactions between keypoints. For each time step $t$, the query $Q$, key $K$, and value $V$ are computed as: 
\begin{equation} 
    Q = h_s^{(t)} W_Q, \quad K = h_s^{(t)} W_K, \quad V = h_s^{(t)} W_V
\end{equation} 

\noindent where $W_Q, W_K, W_V \in \mathbb{R}^{d \times d}$ are learnable weight matrices. The attention output is calculated by taking the inner product of the queries and keys, followed by a softmax operation to produce attention weights: 

\begin{equation} 
    \text{Attention}(Q, K, V) = \text{softmax}\left( \frac{Q \cdot K^\top}{\sqrt{d}} \right) V
\end{equation}

\noindent The spatial encoder processes all keypoints and applies average pooling over the keypoint dimension to produce a spatial feature vector $z_s \in \mathbb{R}^{T \times d}$ for each sequence.

To capture temporal dependencies, the spatial feature vector $z_s$ is combined with a temporal positional embedding $p_t$ to form the input $h_t = z_s \bigoplus p_t$, which is given to the temporal transformer encoder. The temporal encoder, which is similar in structure to the spatial encoder, uses a multi-head self-attention mechanism to learn correlations between time steps. The output of the temporal encoder, $z_t \in \mathbb{R}^{T \times d}$, is pooled along the time dimension to produce a fixed-size representation vector $f \in \mathbb{R}^d$. Finally, $f$ is passed through a fully connected layer followed by a softmax function to generate the final prediction $\hat{y} \in \mathbb{R}^C$.

\subsection{FLAMe algorithm: Attention-driven FL}
\begin{figure}[t]
    \centering
    \includegraphics[width=0.9\linewidth]{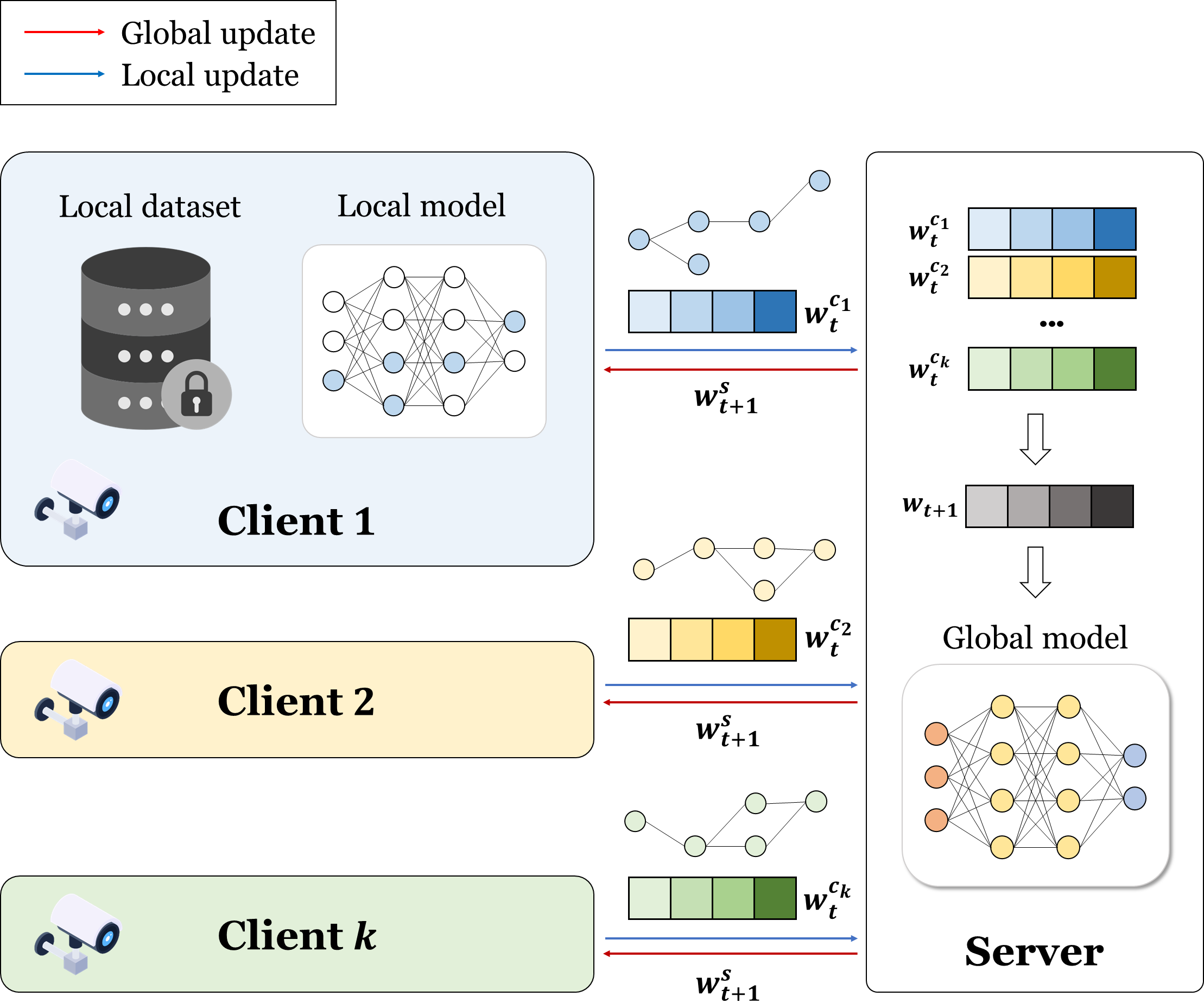}
    \caption{Framework of the proposed FLAMe algorithm. At the beginning of each round $t$, client $c$ downloads the global model weights $w_{t-1}^s$ of the previous round from the server. Each client then trains its local model, selects weights $w_t^{c_k}$ ($k=1, 2, \ldots, K$) for important key points, and uploads them to the server. The server aggregates the weights and calculates the new global model weights $w_t^s$.}
    \label{fig:FLAMe}
\end{figure}

In this section, we describe the process of updating a pedestrian fall detection model using the FLAMe algorithm. As shown in Fig.\ref{fig:FLAMe}, FLAMe algorithm that minimizes communication costs and optimizes learning efficiency by only sending information based on important keypoints to the server. The algorithm is computationally simpler than other widely used FL algorithms such as federated averaging (FedAvg) \cite{mcmahan2017communication}, federated stochastic gradient descent (FedSGD) \cite{mcmahan2017communication}, and federated proximal (FedProx) \cite{li2020federatedoptimizationheterogeneousnetworks}, and offers practical advantages in complex environments in smart cities.

By considering the core encoder structure inside the proposed model, the FLAMe algorithm introduces an aggregation method that effectively reflects spatial information. In addition, it utilizes attention mechanisms to selectively communicate important keypoint information to reduce unnecessary data transmission and improve the performance of the global model. This design is optimized to enable efficient learning even in environments with limited resources.

Algorithm~\ref{algo:flame} presents the pseudocode of the FLAMe algorithm, where the learning process consists as follows: At the first communication round $t=0$, the server initializes the model weights to $\mathbf{w}_0$ and distributes them to each client. Each client trains important keypoints based on its local dataset $D$ and updates its local model by computing the gradient $\nabla \ell(\mathbf{w}_t; b)$ of the loss function $\ell(\mathbf{w}_t; b)$ using its local batch size $B$, learning rate $\eta$, and epoch $E$. This is a critical mechanism to reduce the amount of data transmission and increase learning efficiency.

\begin{algorithm}[tb]
\caption{FLAMe algorithm}
\label{algo:flame}
\textbf{Input}: Communication rounds $T$, number of clients $K$, local batch size $B$, learning rate $\eta$, local epochs $E$ \\
\textbf{Output}: Global model weights $\mathbf{w}_T$

\begin{algorithmic}[1]
\STATE Initialize global model weights $\mathbf{w}_0$ at $t=0$
\FOR{each round $t = 1, \dots, T$}
    \FOR{each client $k \in \{1, 2, \dots, K\}$ \textbf{in parallel}}
        \STATE Client $k$ initializes local model with $\mathbf{w}_t$
        \FOR{each local epoch $e = 1, \dots, E$}
            \FOR{each batch $b \in B$}
                \STATE Compute gradient $\nabla \ell(\mathbf{w}_t; b)$ using local data $D_k$
                \STATE Update local model weights: $\mathbf{w}_t^{c_k} \gets \mathbf{w}_t^{c_k} - \eta \nabla \ell(\mathbf{w}_t; b)$
            \ENDFOR
        \ENDFOR
        \STATE Apply attention mechanism to select important keypoint weights $\mathbf{w}_t^{c_k}$, send to server
    \ENDFOR
    \STATE Server aggregates received weights using:
    \[
    \mathbf{w}_{t+1} = \frac{1}{K} \sum_{k=1}^K \mathbf{w}_t^{c_k}
    \]
    \STATE Update global model weights: $\mathbf{w}_{t+1}$
\ENDFOR
\STATE \textbf{return} Global model weights $\mathbf{w}_T$
\end{algorithmic}
\end{algorithm}

The server aggregates the local weights $\mathbf{w}_t^{c_k}$ ($k=1, 2, \ldots, K$) collected from the clients in a weighted averaging method to obtain new global model weights $\mathbf{w}_{t+1}$. The aggregation process is performed identically to the traditional FedAvg algorithm, only including the weights trained based on the critical keypoints. The resulting global weights is again distributed to all clients, each of which repeats local training based on it.

This learning process is repeated for a total of $T$ rounds, finally resulting in a global model $\mathbf{w}_T$ that reflects the training results of all clients. The FLAMe algorithm is designed to reduce the communication cost and maintain the performance of the global model while maintaining data privacy. The proposed FL-based pedestrian fall detection system utilizes these FLAMe algorithms, providing a practical solution to efficiently process and integrate various video data while ensuring user privacy.

\section{Experiments and Results}
\subsection{Experimental design}
We designed two experiments to evaluate the effectiveness of the proposed system: 1) We utilized a multi-class task scenario with labels for different types of falls to validate the fall detection accuracy of the proposed model. 2) To validate the effectiveness of the FLAMe algorithm, we performed a comparison experiment by implementing the fall detection model with centralized processing and FLAMe-based processing, respectively. By comparing the parameters and model performance of the methods, we comprehensively demonstrated the effectiveness of the proposed approach.

\subsubsection{Dataset}
\begin{figure}[t]
    \centering
    \begin{minipage}[b]{0.32\linewidth}
        \centering
        \includegraphics[width=\linewidth]{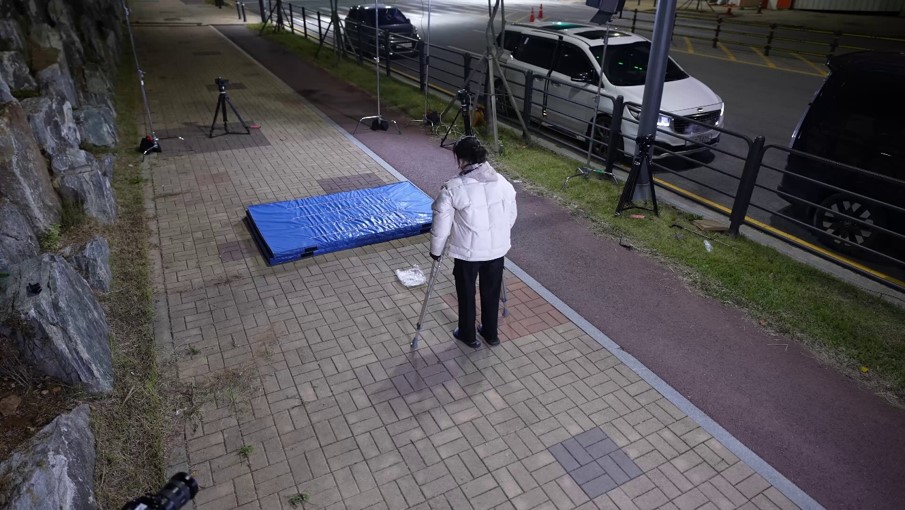}
        \caption*{(a)}
    \end{minipage}
    \hfill
    \begin{minipage}[b]{0.32\linewidth}
        \centering
        \includegraphics[width=\linewidth]{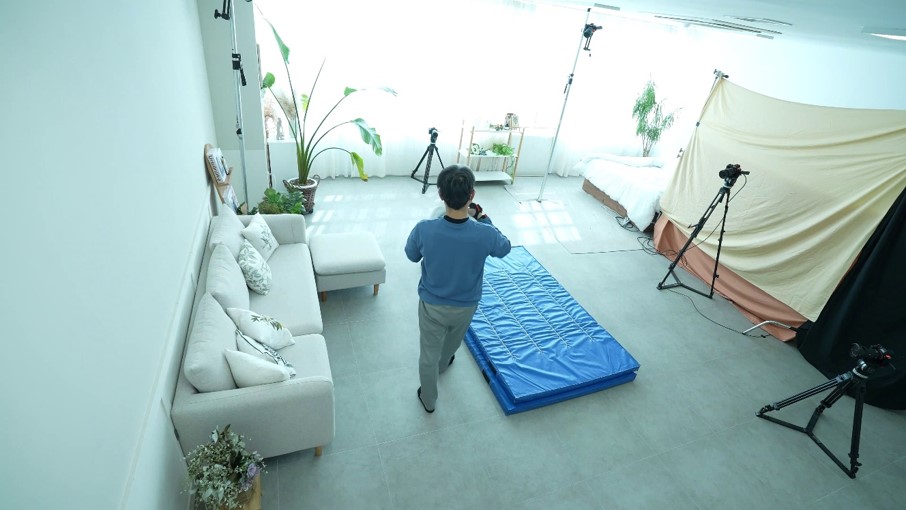}
        \caption*{(b)}
    \end{minipage}
    \hfill
    \begin{minipage}[b]{0.32\linewidth}
        \centering
        \includegraphics[width=\linewidth]{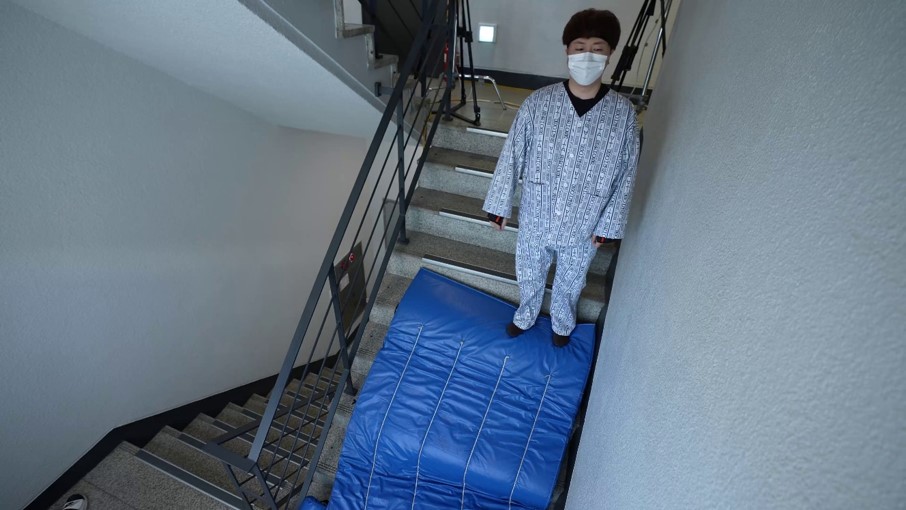}
        \caption*{(c)}
    \end{minipage}

    \vspace{0.1cm} 

    \begin{minipage}[b]{0.32\linewidth}
        \centering
        \includegraphics[width=\linewidth]{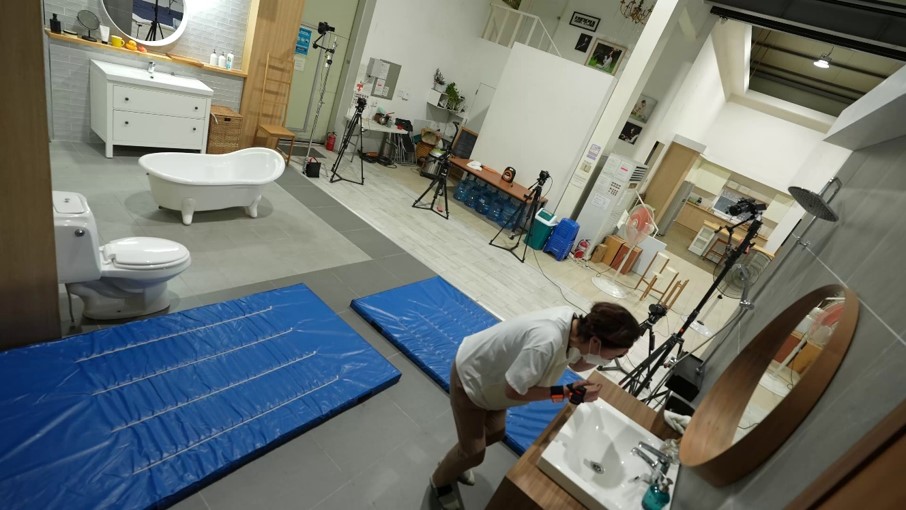}
        \caption*{(d)}
    \end{minipage}
    \hfill
    \begin{minipage}[b]{0.32\linewidth}
        \centering
        \includegraphics[width=\linewidth]{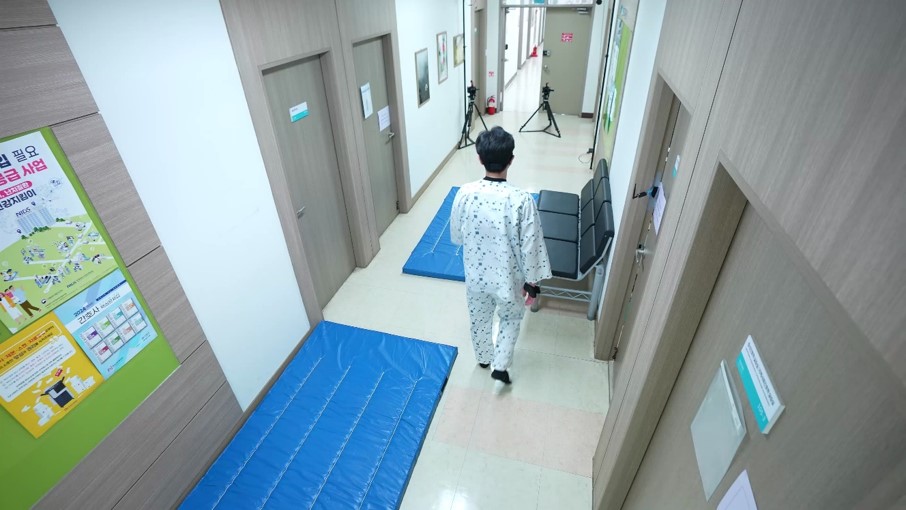}
        \caption*{(e)}
    \end{minipage}
    \hfill
    \begin{minipage}[b]{0.32\linewidth}
        \centering
        \includegraphics[width=\linewidth]{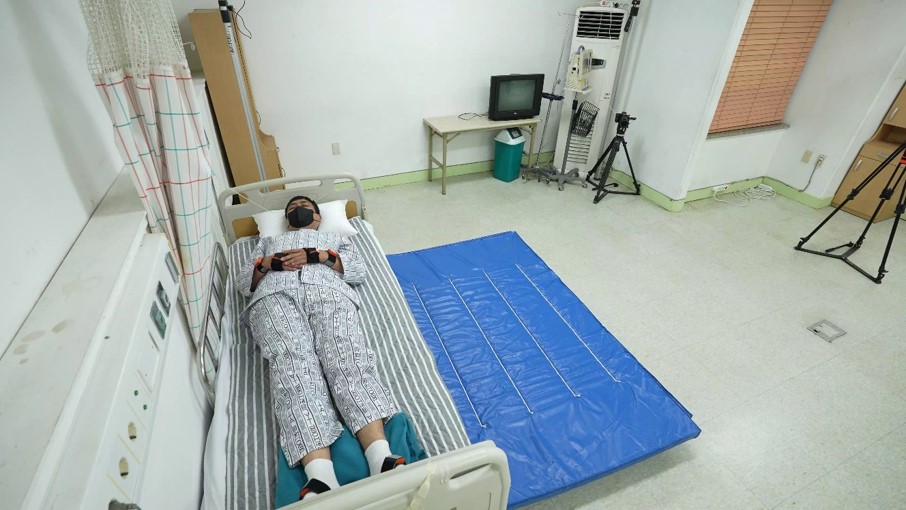}
        \caption*{(f)}
    \end{minipage}
    
    \caption{Examples of videos of the AI-Hub dataset collected in different environments: (a) ground, (b) living room, (c) indoor or outdoor stairs, (d) bathroom, (e) hallway, and (f) hospital.}
    \label{fig:location}
\end{figure}

In our experiments, we validated using the Fall Accident Risk Behavior Video-Sensor Pair Data dataset \cite{AIHubdata} provided by AI-Hub\footnote{This research used dataset from “The Open AI Dataset Project (AI-Hub, S. Korea).” All data information can be accessed through AI-Hub (www.aihub.or.kr).}. It includes video and sensor data, and reflects a variety of environments and scenarios that can occur in real-world fall situations. We only used video data, and the 22,672 video samples are categorized into four classes: no fall, side fall, front fall, and back fall; the details of the class distribution are presented in Tab.\ref{tab:dataset_class}. Each video was captured in a variety of indoor and outdoor environments, including roads, homes, hospitals, welfare facilities, and public places, as shown in Fig.\ref{fig:location}, and utilizes eight angles of the camera to include falls from different perspectives. This configuration of data reflects the variety of environments and scenarios that real-world falls can occur in, and is consistent with the objectives of this study.

\begin{table}[t]
\centering
\caption{Class distribution of the dataset employed in this study}
\label{tab:dataset_class}
\begin{tabular}{@{}ccc@{}}
\toprule
\textbf{ID} & \textbf{Description} & \textbf{\# of videos} \\ \midrule
0                    & Forward Falls        &              7,736        \\
1                    & Sideways Falls       &              3,440        \\
2                    & Backward Falls       &              5,816        \\
3                    & No Falls            &              5,680        \\
\bottomrule
\end{tabular}
\end{table}

\begin{figure}[t]
    \centering
    \begin{minipage}[b]{0.32\linewidth}
        \centering
        \includegraphics[width=\linewidth]{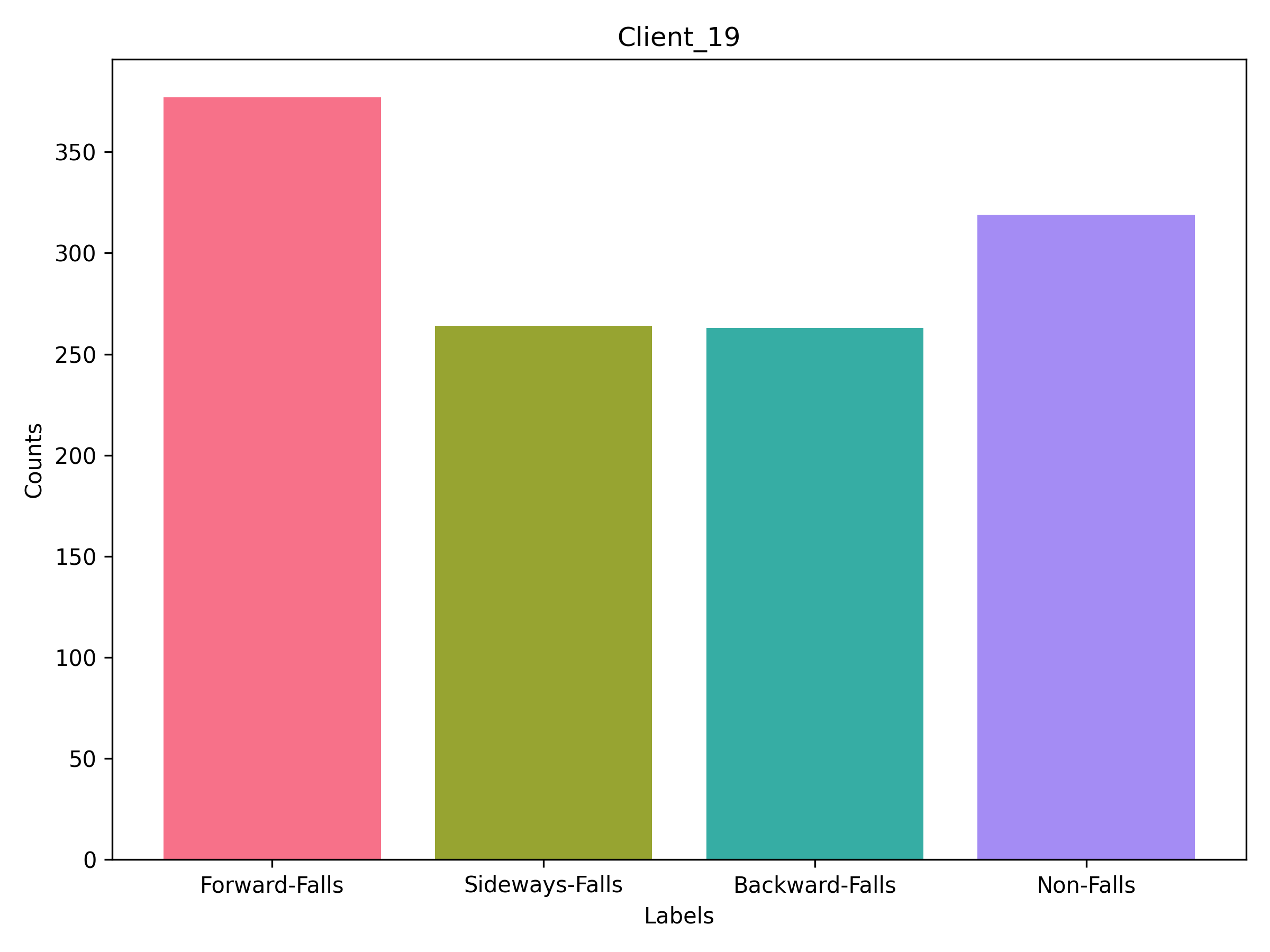}
    \end{minipage}
    \hfill
    \begin{minipage}[b]{0.32\linewidth}
        \centering
        \includegraphics[width=\linewidth]{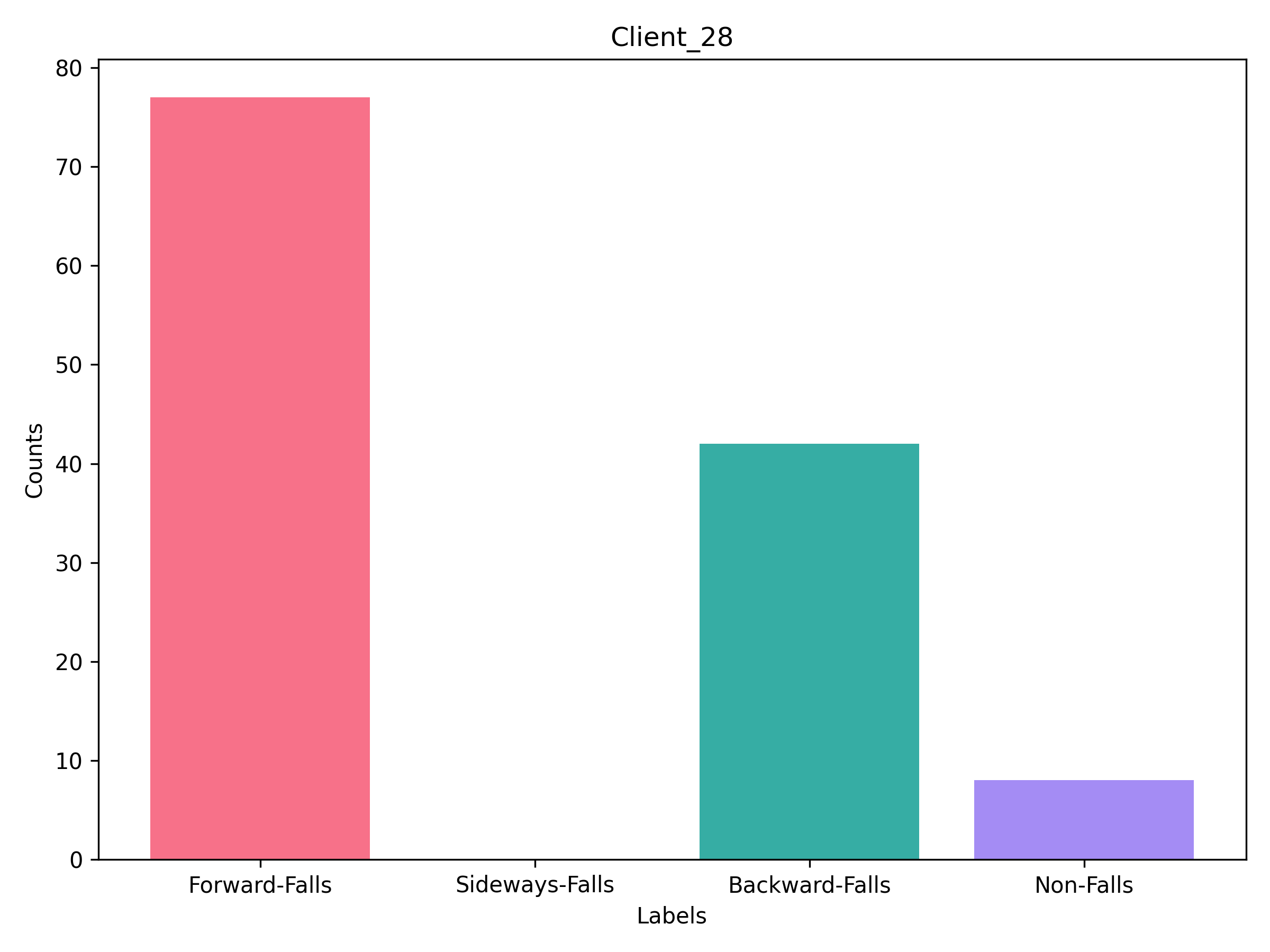}
    \end{minipage}
    \hfill
    \begin{minipage}[b]{0.32\linewidth}
        \centering
        \includegraphics[width=\linewidth]{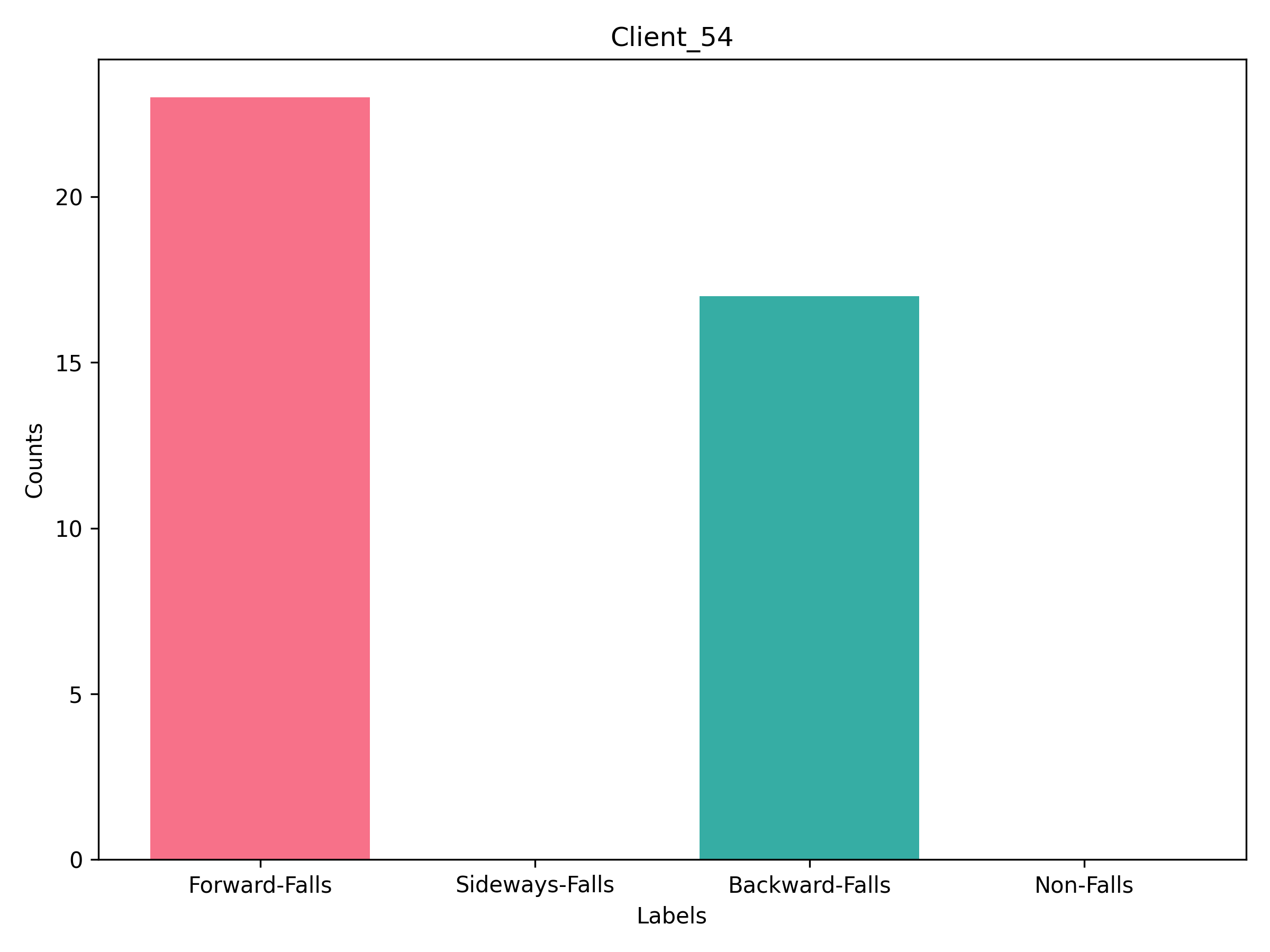}
    \end{minipage}
    \caption{Bar chart example of the distribution of labels for each client}
    \label{fig:distribution}
\end{figure}

In preprocessing, we extracted the video into 3-second segments: 1 second before the start of the fall and 1 second after the end. In addition, we downsampled the frame rate to 15 FPS to clarify the differences between frames, which allowed the model to effectively recognize the subtle body movements of the falling action. The entire dataset was randomly divided into 60:20:20 for training, validation, and testing, respectively. In the FL experiment, we simulated a smart city environment with 56 clients, each representing a CCTV camera in a different location, and each client had its individual local dataset. The data distribution varies from client to client, reflecting a diverse and non-independent and identically distributed (non-IID) data environment with some datasets missing labels and class imbalances. Fig.\ref{fig:distribution} illustrates the distribution of labels for a sample of clients.

\subsubsection{Evaluation metrics}
In our experiments, the accuracies of the proposed fall-detection and baseline models were validated using four commonly used metrics for machine-learning models, namely accuracy, precision, recall, and F1 score. $Accuracy$ represents the proportion of correctly recognized classes. $Precision$ represents the proportion of correctly detected fall events among all predicted fall events, whereas $Recall$ represents the proportion of correctly recognized falls among all actual fall events. Finally, the $F1-score$ is the harmonic mean of precision and recall. In this study, the results for all metrics were expressed in percentages and calculated as follows: 

\begin{equation}
    \text{Accuracy(\%)} = \frac{\text{TP} + \text{TN}}{\text{TP} + \text{TN} + \text{FP} + \text{FN}} \times 100
\end{equation}

\begin{equation}
    \text{Precision(\%)} = \frac{\text{TP}}{\text{TP} + \text{FP}} \times 100
\end{equation}

\begin{equation}
    \text{Recall(\%)} = \frac{\text{TP}}{\text{TP} + \text{FN}} \times 100
\end{equation}

\begin{equation}
    \text{F1-score} = \frac{2\times \text{Precision} \times \text{Recall}}{\text{Precision} + \text{Recall}}
\end{equation} \\
where TP, TN, FP, and FN denote the numbers of true positives, true negatives, false positives, and false negatives, respectively.

\subsubsection{Baselines}
Research on vision-based fall detection using neural networks began in 2012, and the first study to introduce deep learning was in 2014 in \cite{feng2014deep}. Many researchers have since adopted approaches based on conventional CNNs and LSTMs, and these methods have been widely utilized in various studies, such as \cite{alam2022vision, rastogi2022human}. For example, the authors in \citet{gomes2022multi} implemented a real-time fall detection system using 3D-CNN or a combination of 2D-CNN and LSTM, and the authors in \citet{inturi2023novel} achieved high accuracy with a 1D-CNN and LSTM-based model using human pose data as input. These models are still widely used in many studies as they can provide effective fall detection solutions in environments with limited computing resources.

In this study, we compare the performance of the proposed model with the 3D-CNN model proposed in the study of \cite{gomes2022multi}, adopting the structure of the 2D-CNN and LSTM models as the baseline. In addition, we further constructed a GCN-based method that utilizes keypoint information for comparison. Through this comparison, we confirm that the fall detection performance of the proposed model is outperformed by the existing baseline models and focus on demonstrating the effectiveness of the proposed method.

\subsubsection{Hyperparameter}
The hyperparameters used for the proposed FLAMe algorithm and spatio-temporal keypoint transformer are summarized as follows:

\begin{itemize}
    \item Learning Rate ($\eta$): 0.001
    \item Batch Size ($B$): 16
    \item Number of Heads in Multi-Head Attention ($h$): 4
    \item Dimensionality of Embedding ($d$): 64
    \item Epochs per Communication Round ($E$): 3
    \item Communication Rounds ($T$): 100
    \item Keypoint Confidence Threshold: 0.3
    \item Dropout Rate: 0.1
    \item Maximum Sequence Length: 45 frames
\end{itemize}

All experiments were conducted with the above hyperparameter settings, ensuring consistency and reproducibility of the results.

\subsection{Fall-detection performance}
\begin{figure}[t]
    \centering
    \begin{minipage}[b]{0.49\linewidth}
        \centering
        \includegraphics[width=\linewidth]{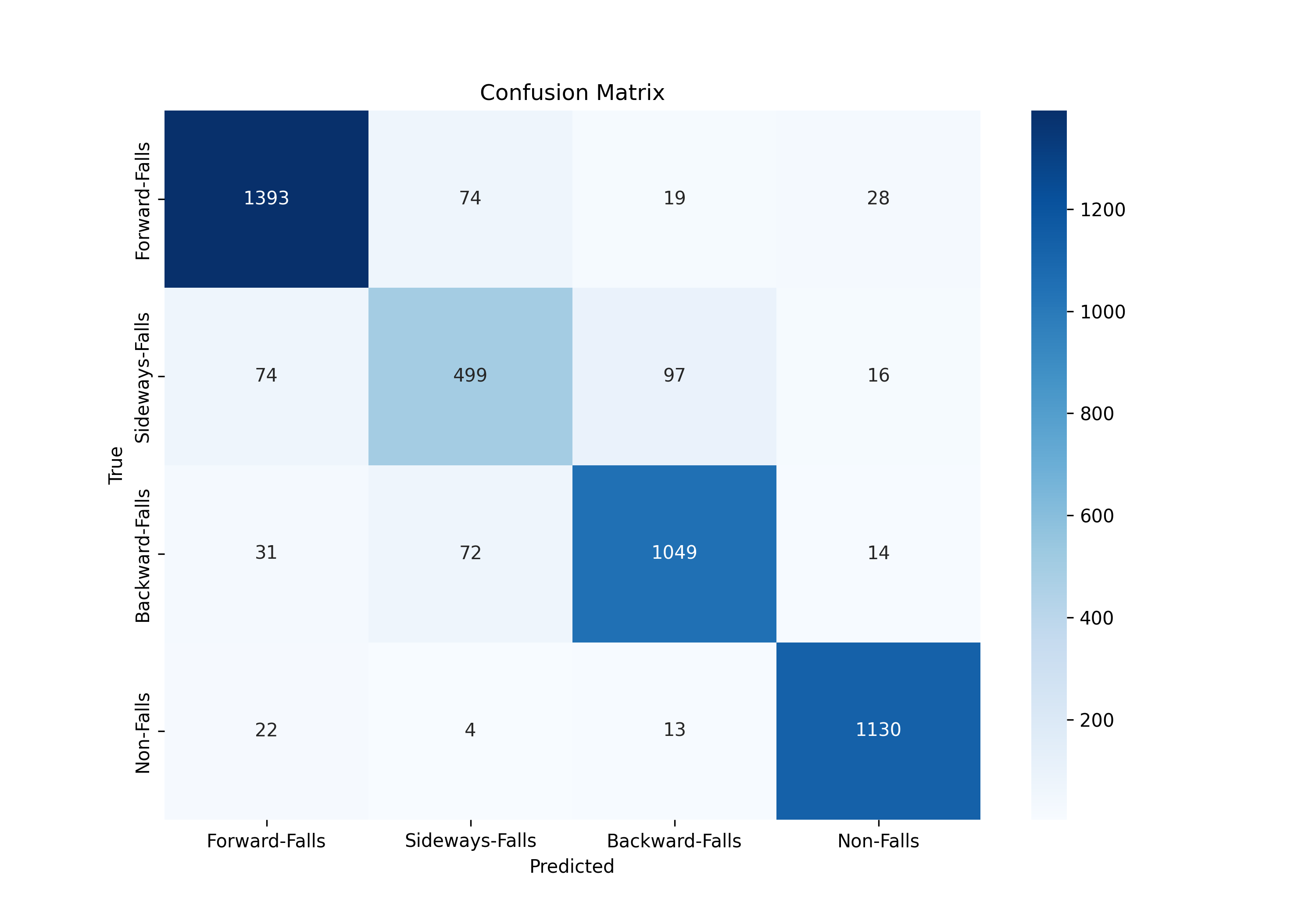}
        \caption*{(a)}
        \label{fig:cnn_lstm}
    \end{minipage}
    \begin{minipage}[b]{0.49\linewidth}
        \centering
        \includegraphics[width=\linewidth]{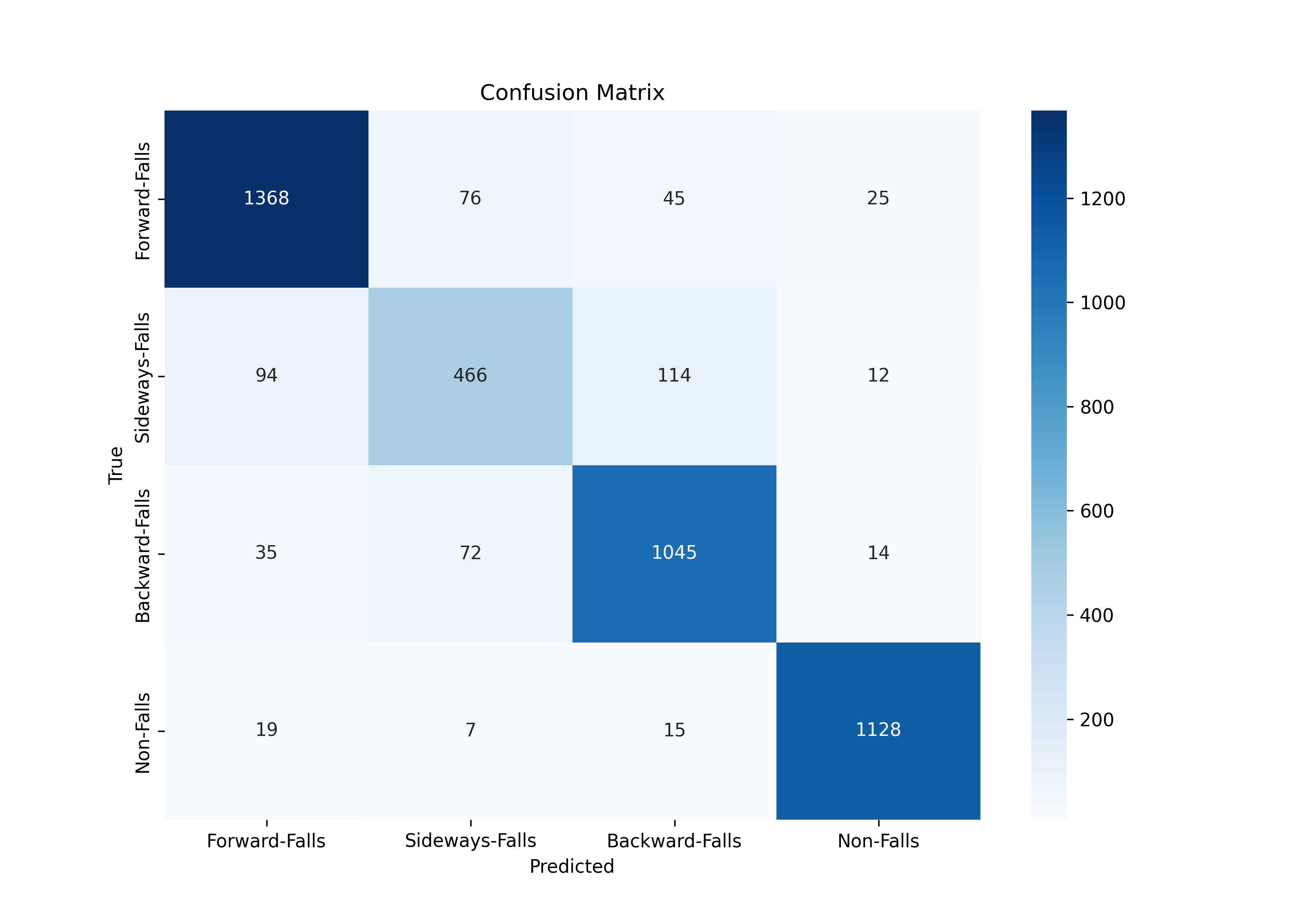}
        \caption*{(b)}
        \label{fig:cnn3d}
    \end{minipage}

    \begin{minipage}[b]{0.49\linewidth}
        \centering
        \includegraphics[width=\linewidth]{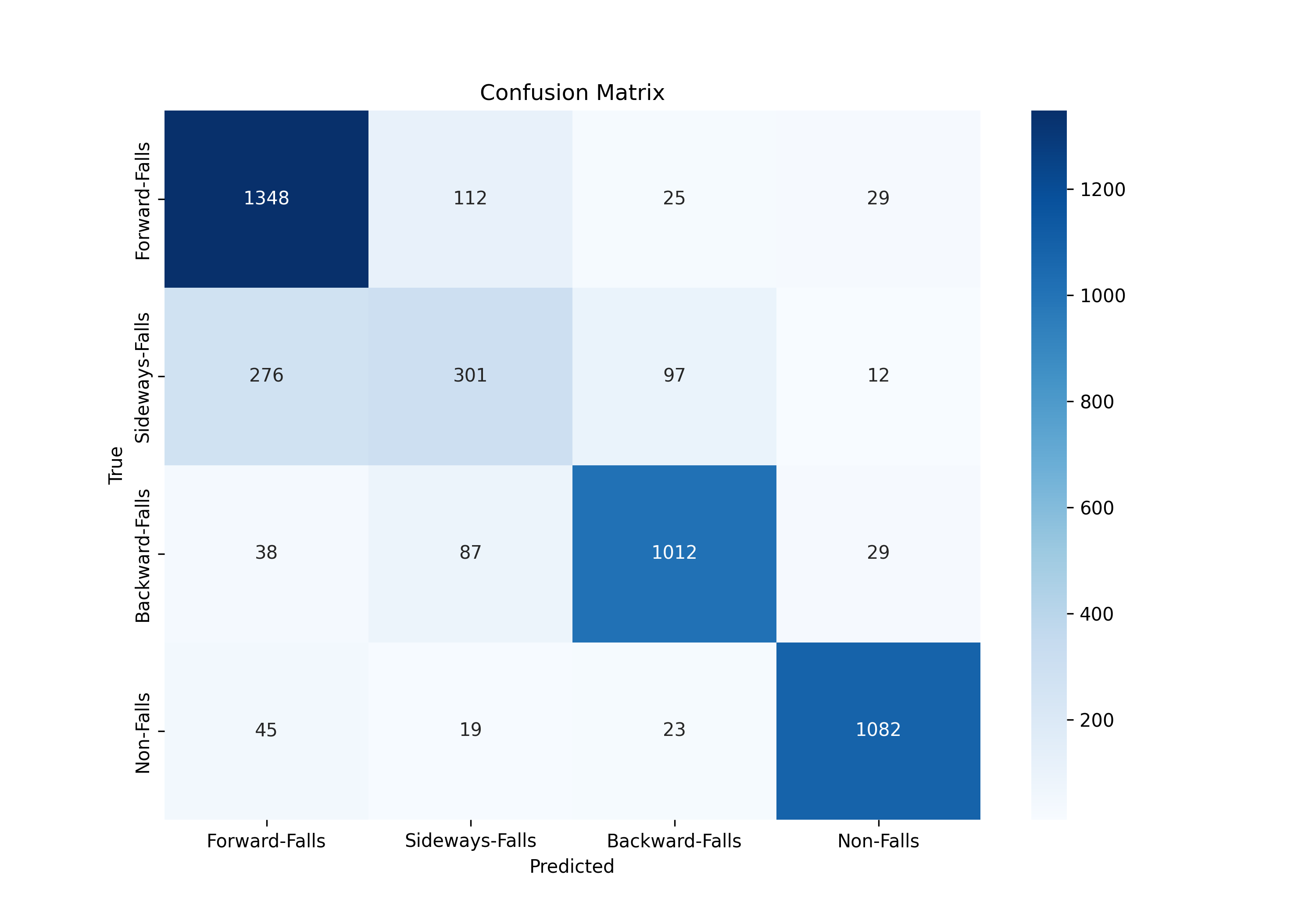}
        \caption*{(c)}
        \label{fig:gcn_lstm}
    \end{minipage}
    \begin{minipage}[b]{0.49\linewidth}
        \centering
        \includegraphics[width=\linewidth]{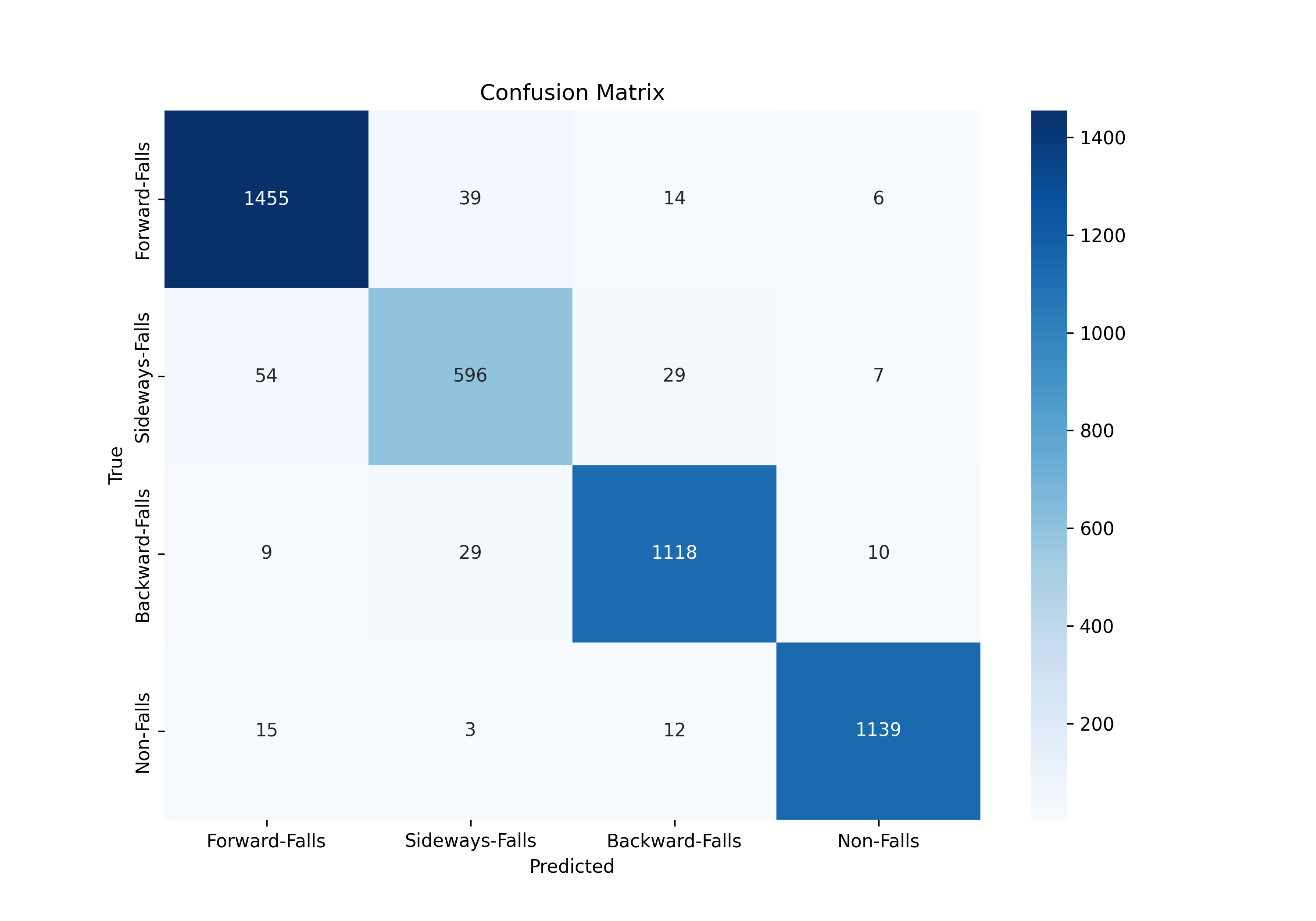}
        \caption*{(d)}
        \label{fig:transformer}
    \end{minipage}

    \caption{Confusion matrices of the (a) 2D-CNN+LSTM, (b) 3D-CNN, (c) GCN+LSTM, and (d) proposed models.}
    \label{fig:confusion_matrix}
\end{figure}

Tab.\ref{tab:fall_detection} shows that the experimental results of the proposed model outperformed the baseline model on all evaluation metrics. The proposed model achieves accuracy of about 94.99\%, precision of about 94.40\%, recall of about 94.08\%, and F1 score of about 94.23\%, which is about 5\% better than the existing highest-performing baseline model, 2D-CNN+LSTM. It uses at least 23 fewer parameters than the CNN+LSTM and 3D-CNN models using RGB images, and achieves comparable performance with less than 50\% fewer parameters compared to the GCN+LSTM approach using the same keypoint data. These results demonstrate that the proposed model is suitable for real-world environments with limited memory resources. Additionally, the confusion matrices in Fig.~\ref{fig:confusion_matrix} illustrate the classification accuracy of each model, demonstrating that the proposed model achieved the lowest error rate among all models.

\begin{table*}[t]
\centering
\caption{Results of the proposed and baseline fall detection models on the AI-Hub dataset}
\label{tab:fall_detection}
\begin{tabular}{@{}ccccccc@{}}
\toprule
\textbf{Model} & \textbf{Accuracy (\%)} & \textbf{Precision (\%)} & \textbf{Recall (\%)} & \textbf{F1-score} & \textbf{Params (M)} \\ \midrule
2D-CNN+LSTM  & 89.77 & 88.17 & 87.84 & 87.99 & 7.47 \\
3D-CNN       & 88.36 & 86.67 & 86.10 & 86.33 & 72.53 \\
GCN+LSTM     & 82.54 & 79.59 & 78.07 & 78.51 & 0.77 \\ \midrule
Ours & \textbf{94.99} & \textbf{94.40} & \textbf{94.08} & \textbf{94.23} & \textbf{0.32} \\ 
\bottomrule
\multicolumn{6}{l}{*Note: M = Million}
\end{tabular}
\end{table*}

\begin{figure}[t]
    \centering
    \includegraphics[width=0.9\linewidth]{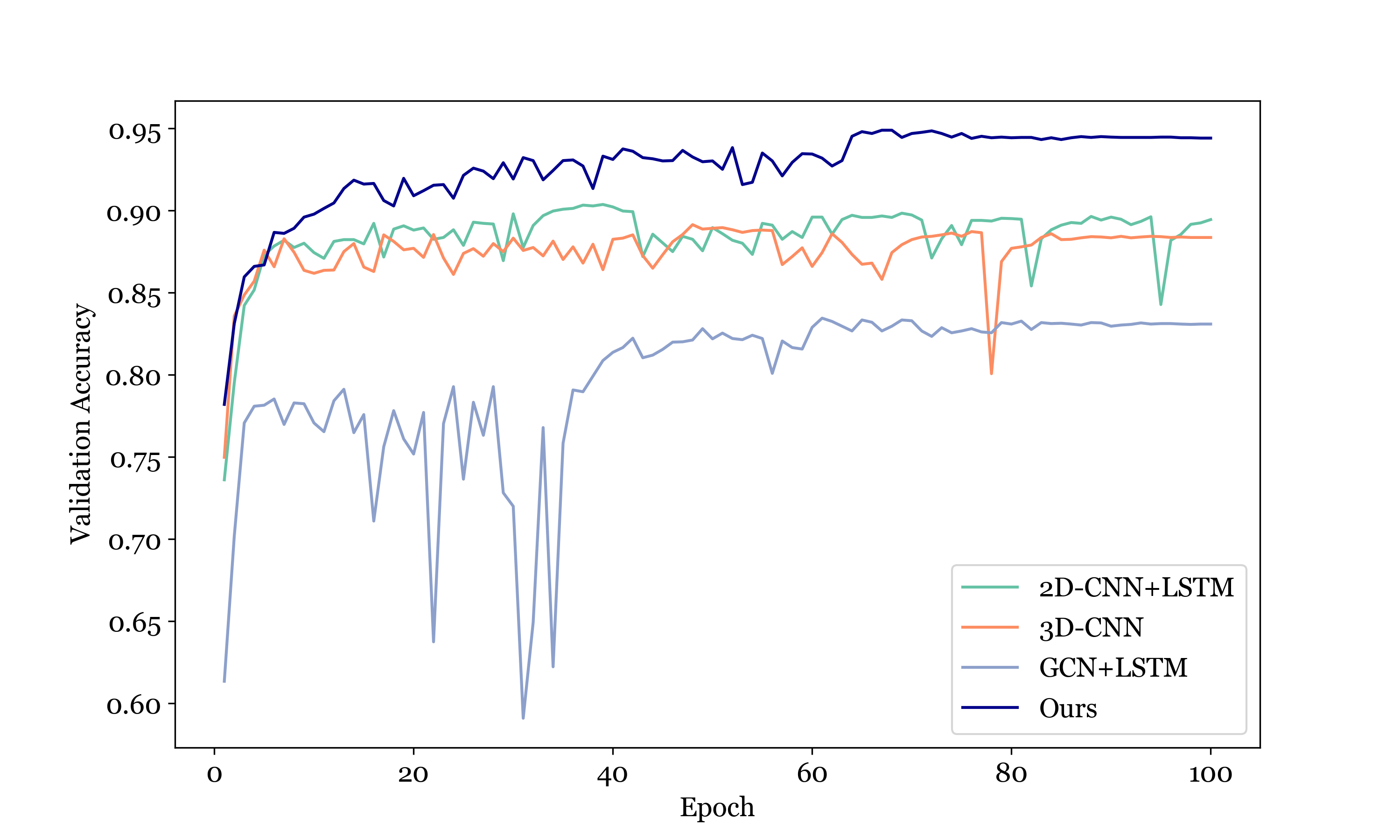}
    \caption{Validation accuracy curves of the proposed and baseline models}
    \label{fig:model_comparison}
\end{figure}

Fig.\ref{fig:model_comparison} visualizes the validation accuracy variation by epoch, which illustrates that the proposed model converged faster and maintained consistently high validation performance. These results emphasize the potential of the proposed model to provide an effective and efficient solution for real-world fall detection systems.

\begin{figure}[t]
    \centering
    \begin{minipage}[b]{0.9\linewidth}
        \centering
        \includegraphics[width=\linewidth]{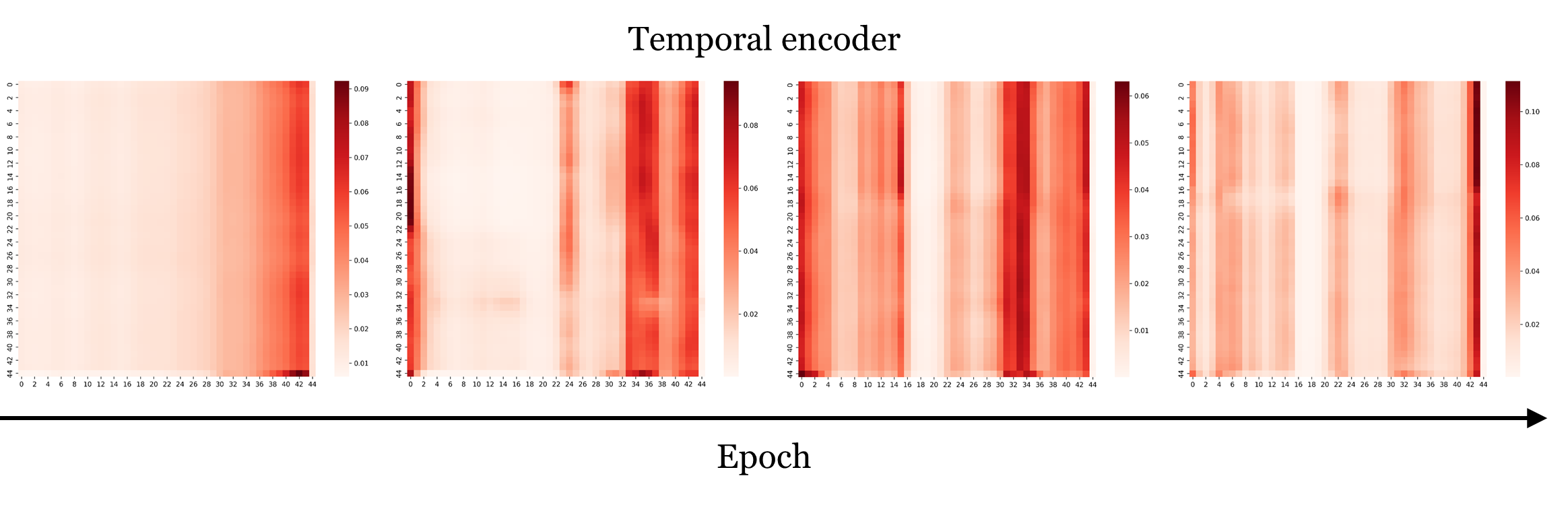}
    \end{minipage}
    \begin{minipage}[b]{0.9\linewidth}
        \centering
        \includegraphics[width=\linewidth]{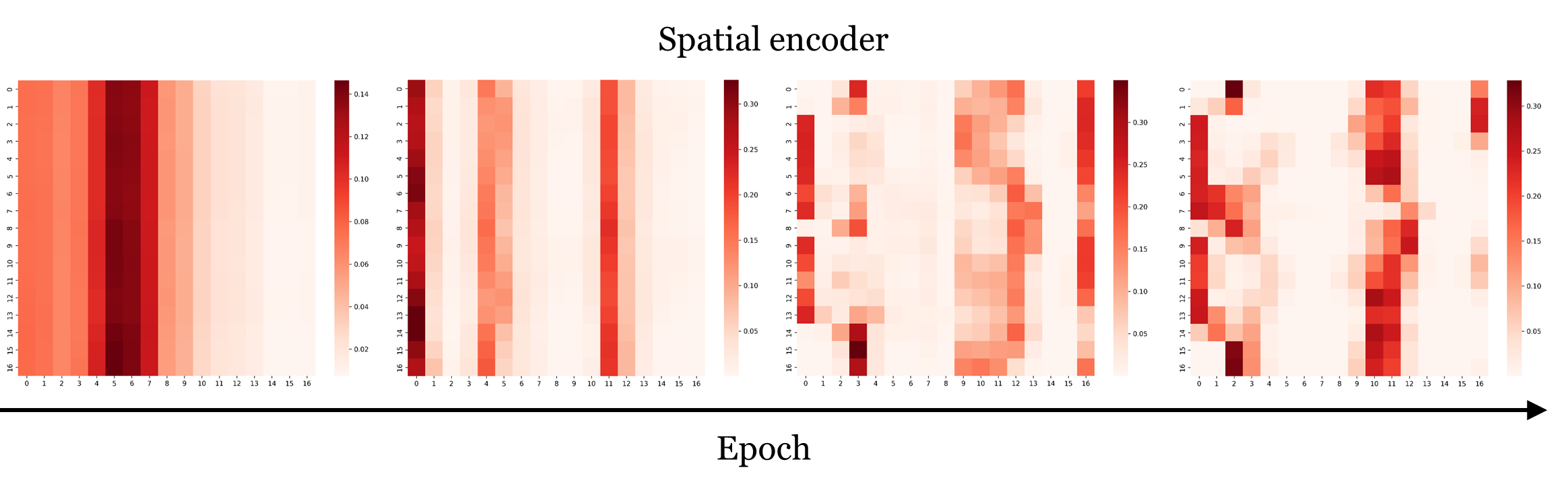}
    \end{minipage}
    \caption{Attention maps of the temporal encoder and spatial encoder as training progresses.}
    \label{fig:attention_map}
\end{figure}

To qualitatively evaluate the performance of the proposed spatio-temporal keypoint transformer model and to check the changes in the spatial and temporal features that the model focused on during training, we visualized the attention map for each epoch. Fig.\ref{fig:attention_map} shows the attention map obtained from the last layer of the spatial and temporal encoder. The temporal encoder reflects the temporal context of the sequence, showing a pattern of evenly distributed attention at the beginning of training, which then gives high weight to the start and end frames of the sequence, especially the last frame.

Furthermore, at the beginning, the spatial encoder evenly distributed attention across multiple keypoints, while as training progressed, it tended to focus on important body parts related to fall detection, such as the face (keypoints 0, 1, 2, and 3), lower body, and right wrist (keypoints 10, 11, and 12). The weights for this important key point information are then reflected as important weights in the global model update during the federated learning process, which helps to improve the learning performance of the global model while reducing the amount of data transmission. As a result, the proposed model effectively learns key information related to fall detection and demonstrates that it can provide high efficiency and performance in real-world environments.

\begin{table*}[!t]
\centering
\caption{Results of centralized, FedAvg, and FLAMe methods}
\label{tab:federated_result}
\begin{tabular}{@{}cccccc@{}}
\toprule
\textbf{Methods} & \textbf{Accuracy (\%)} & \textbf{Precision (\%)} & \textbf{Recall (\%)} & \textbf{F1-score} & \textbf{Transmission parameters (M)} \\ \midrule
Centralized Learning      & 94.99             & 94.40              & 94.08           & 94.23      & -       \\
FedAvg        & 93.08             & 93.23              & 93.08           & 92.76     & 0.32        \\ \midrule
\textbf{FLAMe}        & 94.02             & 92.45              & 92.14           & 91.57      & 0.19       \\ \bottomrule
\multicolumn{6}{l}{*Note: M = Million, -: Not applicable}
\end{tabular}
\end{table*}

\subsection{Performance of the FLAMe algorithm}
We evaluated the efficiency of the proposed FLAMe algorithm in experiments. FLAMe is designed to share only the important weights of the model that each client has trained on its local data to the server, based on the important keypoint information. This can significantly reduce the communication cost while effectively increasing the training performance of the global model. However, previous fall detection experiments have observed that at the beginning of training, the model tends to distribute its attention evenly among all keypoints, and the important keypoints become increasingly obvious as the epoch progresses. To address this challenge, we adopted the approach of sending all keypoint information to the server for the first 30 rounds of the experiment, and selectively sending only weights based on important keypoints in subsequent rounds.

Tab.\ref{tab:federated_result} shows the comparison of the performance of centralized learning, FedAvg, and FLAMe algorithms. The proposed FLAMe algorithm significantly improves communication efficiency by reducing the size of the transmission parameters by about 40\% over centralized learning and FedAvg, while maintaining similar accuracy to FedAvg. FLAMe performed about 1\% lower than centralized learning, which is acceptable in practice considering that FL offers the advantages of keeping user data local, protecting privacy, and learning reliably in a distributed environment.

As a result, FLAMe demonstrates that it can effectively maintain the performance of a global model while reducing unnecessary data transmission by utilizing important keypoint-based information. Therefore, FLAMe provides a practical solution to achieve both high accuracy and reduced communication costs in distributed learning environments, and has demonstrated the potential to be effectively utilized in network- limited environments such as smart cities.

\subsection{Discussions}
This study proposed the FL-based fall detection system using the FLAME algorithm to enhance public safety in smart cities. The proposed system introduces a lightweighted approach to reduce the local computational overhead while achieving the key goals of federated learning: communication efficiency and data privacy protection. In the experiment, the proposed spatio-temporal keypoint-based model outperformed the existing fall detection model in all evaluation metrics, and showed that it maintained high accuracy while significantly reducing memory usage compared to the RGB-based method or the existing keypoint-based model. In addition, the FLAMe algorithm reduced the transmission parameters by about 40\% compared to the existing federated learning algorithm, FedAvg, while showing similar or better performance, maximizing communication efficiency.

The FLAMe-based system showed about 1\% lower accuracy than the centralized learning in terms of performance. This is due to the non-IID characteristics of the data distribution between clients in the FL environment. However, this performance difference is acceptable enough in that the FLAMe algorithm can maintain stable training performance in a distributed environment while reducing communication costs and protecting data privacy. Specifically, compared to FedAvg, FLAMe showed high performance while reducing network resources through efficient data transmission using important key point information. These results demonstrate the practical applicability and high potential of the FLAMe algorithm in distributed environments such as smart cities.

\section{Conclusion}
In this study, a FL-based fall detection system using the FLAME algorithm was proposed to enhance public safety in smart cities. FLAMe performs local training using important key point information and only sends key weights to the server, reducing communication costs and protecting data privacy. 

As a result of the experiment, the lightweight model, which uses about 190,000 transmission parameters, achieved high accuracy (94.02\%), proving its efficiency in reducing communication costs by about 40\% compared to FedAvg while maintaining performance similar to centralized learning. Specifically, the proposed lightweight keypoint transformer model reduces the local computation overhead, further improving the computational efficiency compared to existing FL algorithms. This design balances communication efficiency and local computation optimization, enhancing the practical contribution of the FLAMe algorithm. 

The proposed FLAMe algorithm reduces network bottlenecks and data privacy issues, and has demonstrated high applicability in smart city environments. Future work will involve using more various datasets, theoretical convergence analysis, and comparative experiments with various FL algorithms to extend FLAMe to various environments and public safety applications.


\section{Acknowledgments}
This work was supported by Institute of Information \& communications Technology Planning \& Evaluation (IITP) grant funded by the Korea government(MSIT) (No.RS-2024-00459703, Development of next-generation AI integrated mobility simulation and prediction/application technologies for metropolitan cities)

\bibliography{aaai25}

\end{document}